\documentclass[10pt,twocolumn,letterpaper]{article}

\usepackage{cvpr}              

\usepackage[dvipsnames]{xcolor}

\definecolor{cvprblue}{rgb}{0.21,0.49,0.74}
\usepackage[pagebackref,breaklinks,colorlinks,citecolor=cvprblue]{hyperref}

\usepackage{amsmath}
\usepackage{graphicx} 
\usepackage{subcaption} 
\usepackage{float}

\def\paperID{*****} 
\def\confName{CVPR}
\def\confYear{2024}

\title{Parameterized Brushstroke Style Transfer\thanks{This paper presents a PyTorch implementation of the CVPR 2021 method ``Rethinking Style Transfer: From Pixels to Parameterized Brushstrokes.''}}

\author{
Uma Maheswara R Meleti\\
Clemson University\\
Clemson, SC, USA\\
{\tt\small umeleti@clemson.edu}
\and
Siyu Huang\\
Clemson University\\
Clemson, SC, USA\\
{\tt\small siyuh@clemson.edu}
}

\begin{document}
\maketitle
\begin{abstract}

Computer Vision-based Style Transfer techniques have been used for many years to represent artistic 
style. However, most contemporary methods have been restricted to the pixel domain; in other words, the style transfer approach has been modifying the image pixels to incorporate artistic style. However, real artistic work is made of brush strokes with different colors on a canvas. Pixel-based approaches are unnatural for representing these images. Hence, this paper discusses a style transfer method that represents the image in the brush stroke domain instead of the RGB domain, which has better visual improvement over pixel-based methods.

{\href{https://maheshmeleti.github.io/param-brushstroke/}{https://maheshmeleti.github.io/param-brushstroke/}}

\end{abstract}    

\section{Introduction}
\label{sec:intro}

Style transfer has been an essential technique in computer vision, extensively researched for decades. It is now widely used in various applications such as animated content, fashion design, mobile app photo filters, etc. The technique enables the transformation of images by applying the visual style of one image (the style image) to another (the content image). The groundbreaking work of  Gatys et al. \cite{gaty}, which introduced the concepts of content and style loss, has paved the way for many subsequent developments in the field. Most works before Gaty's method and subsequent methods operate within the pixel domain,  where image pixels are manipulated to incorporate artistic styles.

However, all these contemporary methods fall short of mimicking the actual artistic work characterized by brush strokes and texture on a canvas. Pixel-based methods often produce images lacking the natural flow and feel of brush strokes.

In contrast to pixel-based methods, this paper introduces a novel approach that performs style transfer in the brush stroke domain rather than the RGB pixel domain. By parameterizing brush strokes and defining their location, color, width, and shape, our method aims to generate visually closer results to hand-painted artwork. This representation better preserves the artistic integrity of the original style, providing a more realistic and visually appealing output.

The brush strokes are modeled as a bezier curve with color, width, and location on the canvas, and these are optimized to produce stylized outputs. We have introduced a renderer that maps brush strokes into pixel values on canvas. The render is made differentiable, allowing gradients to backpropagate and optimize the brush stroke parameters for producing stylized outputs.

\section{Related Work}
\label{sec:related work}

In the earlier works on style transfer, Efros and Freeman \cite{Efros01} performed texture synthesis and transfer using image quilting, an algorithm that stitches small blocks of texture from an image in a way that resembles the input image.   Hertzmann et al. \cite{10.1145/3596711.3596770} proposed a method that uses a pair of images—one being a filtered version of the other—to learn a transformation filter. Given an image pair, \(A\) and \(A'\), the algorithm learns how pixel neighborhoods (or textures) in image \(A\) correspond to those in image \(A'\). This learned filter can then be applied to a new image, \(B\), effectively transferring the learned texture transformation to the new image.

The most seminal work on style transfer was proposed by Gatys et al. in their paper \textit{Image Style Transfer Using Convolutional Neural Networks} \cite{gaty}. In this approach, the content image and style image are used as inputs to generate a new image that retains the content of the content image while adopting the style of the style image. The method introduces a novel loss function that separates content and style components, utilizing a pre-trained neural network to extract content and style representations from the given images. 

\begin{figure*}[t]  
    \centering
    \includegraphics[width=\textwidth, height=0.25\textheight]{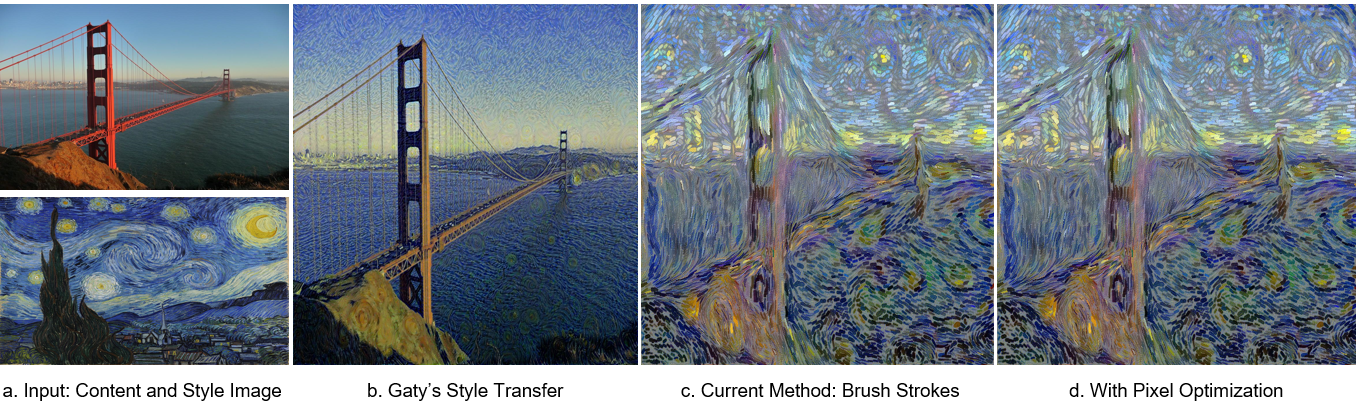}  
    \caption{B represents the output of Gatys' style transfer method, while C is the result of our proposed approach. Visually, our method produces an image that more closely resembles artistic brush strokes compared to Gatys'. D is the output after pixel optimization, where the brush strokes are blended, resulting in a more cohesive and refined representation.
}
    \label{fig:imageA}
\end{figure*}

Later, Huang and Belongie improved style transfer using feed-forward networks for arbitrary style transfer by introducing Adaptive Instance Normalization (AdaIN) \cite{huang2017arbitrarystyletransferrealtime}. AdaIN is a layer (or block) that aligns the mean and variance of the content features with those of the style features. This block transforms the extracted features of content and style images from a fixed VGG-19 network, and a decoder is used to invert the AdaIN output back into the image space.

Traditional content loss measures how well the structure of the content is preserved in the stylized output. However, this approach doesn't always account for the effects of stylization, which can sometimes result in a blurred or less accurate rendering, particularly in high-resolution images. To address this, Sanakoyeu et al. and Kotovenko et al. introduced style-aware content losses \cite{sanakoyeu2018styleawarecontentlossrealtime} to enhance stylization quality. These losses modify traditional content loss by considering the style features, which better balance content preservation and stylization.

In addition to pixel-based approaches, some methods focus on rendering brush strokes. Early works in this area include an interactive technique by Haeberli \cite{10.1145/97879.97902}, where a program follows the cursor across the canvas, samples the source image at each point to obtain a color, and then paint a brush stroke of that color. Hertzmann \cite{10.1145/280814.280951} extended this approach with an automated algorithm that takes a source image and a list of brush sizes, then paints multiple layers—one for each brush size—on a canvas to recreate the source image with a hand-painted effect. In contrast to these stroke-based rendering methods, some attempts focus on detecting and extracting brush strokes from paintings. POET \cite{POET} uses classical computer vision techniques, such as circular filters and orientation phase extractors, to automatically extract brush strokes. Jia Li et al. \cite{6042878} employ edge detection and clustering-based segmentation methods to extract brush strokes, enabling comparisons between artistic works and other artists' styles.

Even though there has been a significant amount of work on artistic style transfer, the authenticity of brush strokes in paintings is often not preserved on the canvas during stylization. In contrast to the above-mentioned methods, the proposed method utilizes parameterized brush strokes to retain the artwork's original texture and stroke characteristics. Using these parameterized brush strokes, the method aims to better preserve the artistic integrity and hand-painted quality of the source image while applying the desired style.

\section{Approach}
\label{sec: Approach}

The proposed method is inspired by the iterative style transfer approach of Gatys et al. \cite{gaty}, where a white noise image is optimized via gradient descent using content and style loss (see Section \ref{sec: loss_function}) to stylize the content image. In the proposed approach, the optimization process begins by initializing a neural network with
N brush stroke parameters — location, color, width, and curve shape, where N represents the desired number of brush strokes to fill the stylized canvas.

These parameters are iteratively optimized using the same content and style losses (as discussed in Section \ref{sec: loss_function}), ensuring the resulting canvas captures the content of the original image while applying the style of the reference image. The canvas is generated by a differentiable renderer, which takes the optimized brush stroke parameters and produces an RGB image. The brush strokes in this image are guided by their respective locations, widths, and colors. Further details on the differentiable renderer can be found in Section \ref{sec: diff_render}.

The shape of each brush stroke is modeled using a Bézier curve, as shown in eq. \ref{eq:bezier_curve}:

\begin{equation} B(t) = (1 - t)^2 P_0 + 2(1 - t)t P_1 + t^2 P_2 \quad \text{for} \quad t \in [0, 1]. \label{eq:bezier_curve} \end{equation}

\begin{figure*}[t]  
    \centering
    \includegraphics[width=\textwidth, height=0.30\textheight]{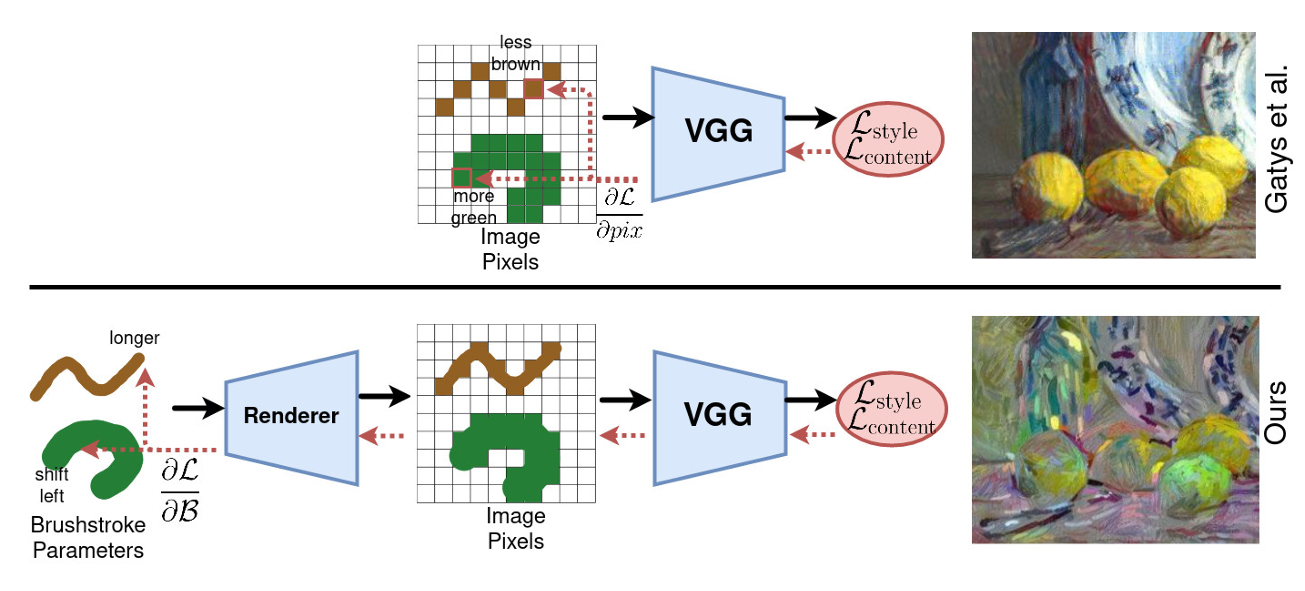}  
    \caption{The top row shows the results of Gatys' style transfer approach, while the bottom row displays the output of the proposed method. In our approach, parameterized brush strokes are passed through a differentiable renderer, which maps them onto the canvas. The content and style losses are then calculated, and gradients are backpropagated through the renderer to optimize the brush strokes. The image is taken from \cite{kotovenko2021rethinkingstyletransferpixels}}
    \label{fig:imageD}
\end{figure*}

After the brush stroke optimization is completed, a pixel-level optimization step is applied, akin to Gatys et al.'s style transfer method. This step blends the brush strokes and adds finer texture details to the final image. Both outputs are presented in the experiments section \ref{sec: experiments}.

\subsection{Differentiable Renderer}
\label{sec: diff_render}

A differentiable renderer maps a collection of brushstrokes (trained) parameterized by location, shape, width and color into pixel values on a canvas. A renderer can be described as a function:

\[
R: \mathbb{R}^{N \times F} \rightarrow \mathbb{R}^{H \times W \times 3}
\]

Where N is the number of brush strokes, F is the number of brushstroke parameters, which is 12 - location: 2, shape: 6 ($P_{0}$ : 2, $P_{1}$ : 2, $P_{2}$ : 2), width : 1, color : 3.

Let's try to understand how the rendering works with a simple example of rendering a flat disk parameterized by its color \(C\), radius \(R\), and location (1, 1, and 2 scalars). We assume our canvas is grayscale, which is just a 2D matrix of pixel values (this can be generalized to RGB space). Our task is to decide whether each pixel belongs to the disk; we compute the L2 distance matrix \(D\) from the disk's center and mask all the values with the color \(C\) where \(D < R\). If \(I_1\) and \(I_2\) are two images obtained for two disks using this procedure, we can get the final image as \(I_1 + I_2\), assuming the disks do not overlap.

If the regions overlap, we compute an assignment matrix \( A \), where the assignment is made to a particular disk if the distance is less than the other disk's. If there are two disks with distance matrices \( D_1 \) and \( D_2 \), we define \( A := \{ 1 \text{ if } D_1 \leq D_2, 0 \text{ otherwise} \} \). This operation can be generalized to \( N \) objects.

\[
A(i,j,n) :=
\begin{cases} 
1 & \text{if } D_n(i,j) < D_k(i,j) \, \forall k \neq n, \\
0 & \text{otherwise}.
\end{cases}
\]

The final image for N discs $I_{1}$, \dots, $I_{N}$ is computed as 

\[
I(i, j) := \sum_{n=1}^{N} I_n(i, j) \cdot A(i,j,n)
\]

To render the disk, we calculate distance matrices, which form the cornerstone of the rendering process. The same principle applies to rendering brushstrokes (or Bézier curves): we compute the distance matrix \( D_B \) by point sampling \( S \) equidistant points \( p_1, \dots, p_S \) along the curve and mask the values that are less than the brushstroke's width.

For practical implementation, the renderer must be differentiable. However, the masking and assignment operations are discontinuous, making them non-differentiable. To resolve this, we make both operations continuous by using a sigmoid function for masking and a softmax function with high temperature for assignment. Additionally, computing the distances between each brush stroke and every pixel is computationally expensive and redundant, as each brush stroke only affects nearby areas of the canvas. Therefore, the distance computation is limited to the K nearest brush strokes.

\subsection{Loss Functions}
\label{sec: loss_function}

Content and style losses are essential for neural style transfer, guiding the balance between preserving the content image's structure and applying the style image's artistic features.

\textbf{Content Loss:}
The content loss ensures that the generated image \(G\) retains the structure of the content image \(C\). It is computed by comparing the feature maps \( \phi_l \) of the content and generated images at a specific layer \( l \) of a pre-trained network (e.g., VGG-19):

\begin{equation}
\mathcal{L}_{\text{content}}(C, G) = \frac{1}{2} \sum_{i,j} \left( \phi_l(G)_{ij} - \phi_l(C)_{ij} \right)^2
\end{equation}

\textbf{Style Loss:}
The style loss measures how well the generated image captures the style image \(S\). It uses the Gram matrix \( G_l(x) \) of feature maps to compute correlations between filter responses. The style loss is defined as:

\begin{equation}
\mathcal{L}_{\text{style}}(S, G) = \sum_l w_l \cdot \frac{1}{4N_l^2 M_l^2} \sum_{i,j} \left( G_l(G)_{ij} - G_l(S)_{ij} \right)^2
\end{equation}

where \( N_l \) and \( M_l \) are the number of filters and the spatial size of the feature maps at layer \( l \), and \( w_l \) is the weight for each layer.

\textbf{Total Loss:}
The total loss \( \mathcal{L}_{\text{total}} \) is a weighted sum of the content and style losses:

\begin{equation}
\mathcal{L}_{\text{total}} = \alpha \mathcal{L}_{\text{content}} + \beta \mathcal{L}_{\text{style}}
\end{equation}

where \( \alpha \) and \( \beta \) control the trade-off between content preservation and style transfer.

\section{Experiments and Results}
\label{sec: experiments}

The model is trained on Nvida-a100 GPUs using the pytorch framework. Optimizing 5000 brush strokes with 10 samples per curve took around 138 seconds.

\begin{figure}[htbp]
    \centering
    \includegraphics[width=0.45\textwidth]{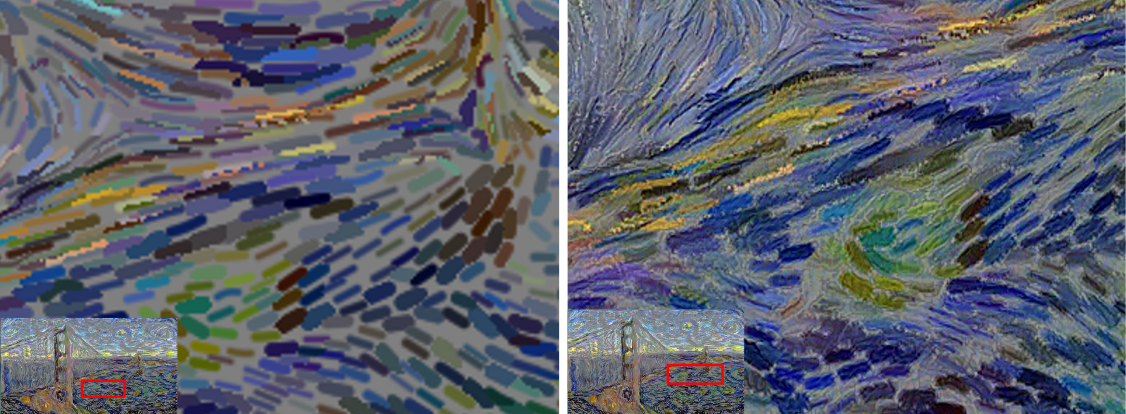}
    \caption{Zoomed view of brush strokes with pixel optimization.}
    \label{fig:imageB}
\end{figure}

Figure \ref{fig:imageA} presents the results from both methods, where our approach yields visually superior representations that are artistically closer to hand-painted works. Figure \ref{fig:imageB} is the zoomed images highlighting the brush strokes generated by our method, followed by the pixel optimization applied to enhance the output.

The pixel-optimized output blends the brush strokes, producing a realistic image that mimics the texture and look of a painting on canvas.

\begin{figure}[htbp]
    \centering
    \includegraphics[width=0.45\textwidth]{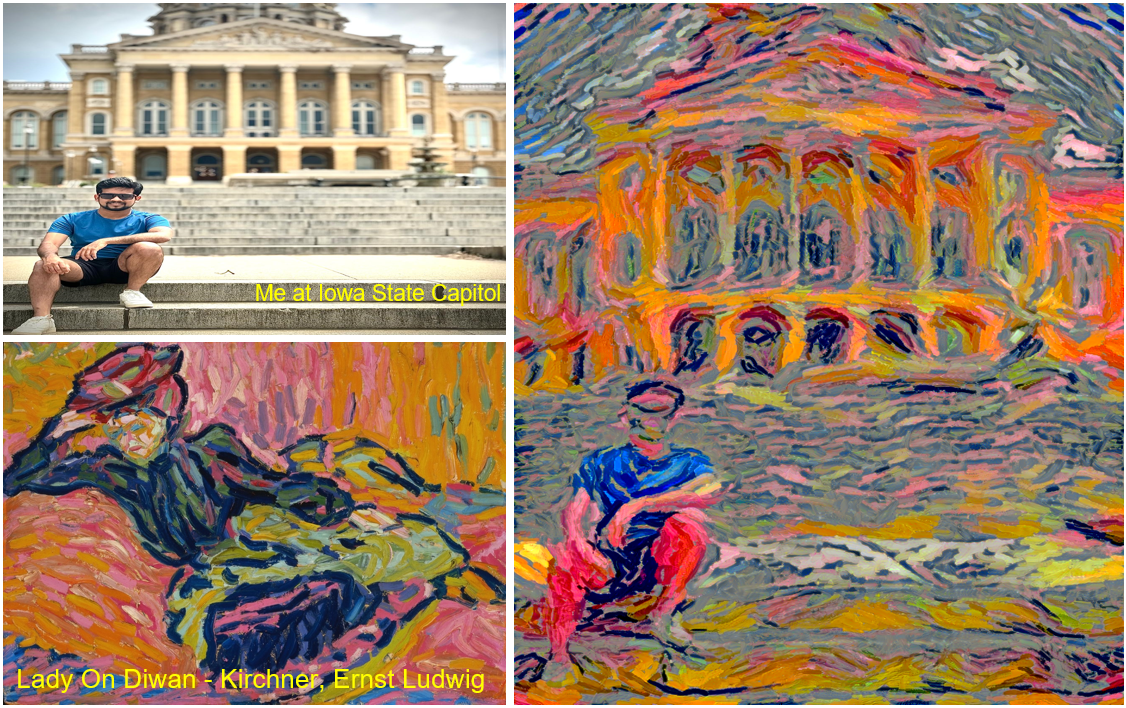}
    \caption{Style transfer applied to a human image.}
    \label{fig:imageC}
\end{figure}

Figure \ref{fig:imageC} is the result of applying our algorithm to a human image. While the algorithm effectively captures the brushstroke representation, finer details, such as facial features, are lost.

 This indicates that the current method struggles to preserve intricate details, suggesting the need for improvements in handling high-frequency content.

\section{Conclusions}
\label{sec: Conclusions}

In this paper, we proposed a new approach for representing artistic style by switching from pixel-based methods to parameterized brush strokes for more natural representations. An explicit rendering mechanism is implemented that can be applied beyond style transfer.

Even though we obtained a brush stroke-based representation, qualitative results indicate some fine details of the content are missing. Implementing CNN-based feed-forward architectures can help mitigate these issues, as the convolutional inductive biases in CNNs effectively preserve rich information on image attributes. By leveraging the hierarchical feature extraction capabilities of CNNs, we can better capture subtle textures and fine-grained information.

Integrating image and text-based methods such as CLIP \cite{clip} can significantly advance research by enabling more sophisticated image editing techniques, offering users finer control over the generated results to better align with their specific goals with the help of language.

\newpage  
{
    \small
    \bibliographystyle{ieeenat_fullname}
    \bibliography{main}
}


\end{document}